\documentclass[times, preprint, 10pt]{elsarticle}
\usepackage{amssymb}
\usepackage{amsmath}
\usepackage{natbib}   %[numbers]
\usepackage{graphicx}%
\usepackage{multirow}%
\usepackage{amsmath,amssymb,amsfonts}%
\usepackage{amsthm}%
\usepackage{mathrsfs}%
\usepackage{xcolor}%
\usepackage{textcomp}%
\usepackage{manyfoot}%
\usepackage{booktabs}%
\usepackage{array}
\usepackage{algorithm}%
\usepackage{algorithmicx}%
\usepackage{algpseudocode}%
\usepackage{listings}%
\usepackage{xspace}
\usepackage{bm}
\usepackage{bbding}
\usepackage{tabu}
\usepackage{url}
\usepackage{xr-hyper}
\usepackage[hidelinks]{hyperref}
\makeatletter
\DeclareRobustCommand\onedot{\futurelet\@let@token\@onedot}
\def\@onedot{\ifx\@let@token.\else.\null\fi\xspace}

\def\etal{\emph{et al}\onedot}
\makeatother

\definecolor{baseline}{rgb}{0.8, 0.8, 0.8}

\newcommand{\revise}[1]{{\color{black}#1}}

\journal{Pattern Recognition}

\begin{document}

\begin{frontmatter}



% \title{}
\title{Frequency-guided adaptive computation for efficient single-image super-resolution}


% \author{} %% Author name
% \author[1]{\fnm{Hannan}\sur{Lu}}\email{hannanlu@hit.edu.cn}
% \equalcont{Equal contributions.}
% \author[2]{\fnm{Xinyu} \sur{Zhang}}\email{zhangxinyu14@baidu.com}
% \equalcont{Equal contributions.}
% \author[3]{\fnm{Zhi} \sur{Tian}}\email{zhi.tian@outlook.com}
% \author[2]{\fnm{Xiangbo} \sur{Su}}\email{suxiangbo@baidu.com}
% \author[2]{\fnm{Gang} \sur{Zhang}}\email{zhanggang03@baidu.com}
% \author*[1]{\fnm{Wangmeng} \sur{Zuo}}\email{wmzuo@hit.edu.cn}
% \author[2]{\fnm{Jingdong} \sur{Wang}}\email{wangjingdong@baidu.com}

\author[1]{Wei Shang}  %[orcid=0000-0001-6039-6041]
\affiliation[1]{organization={Singapore Management University},
	%                addressline={Jawahar Nagar}, 
	%                city={Trivandrum},
	%               citysep={}, % Uncomment if no comma needed between city and postcode
	%                postcode={695013}, 
	%                state={Kerala},
	country={Singapore}}
\author[2]{Wanying Zhang}   %[orcid=0009-0009-3618-0895]
\affiliation[2]{organization={Harbin Institute of Technology},
	%                addressline={Street 15}, 
	%                city={Jabaldesh},
	%                postcode={825001}, 
	%                state={Orissa}, 
	country={China}}

\author[3]{Wenwen Li \corref{cor1}}  
\affiliation[3]{organization={Tianjin Medical University},
	%                addressline={Street 15}, 
	%                city={Jabaldesh},
	%                postcode={825001}, 
	%                state={Orissa}, 
	country={China}}

\author[2]{Xiaohe Wu} 	
\author[4]{Huaming Wu}  
\affiliation[4]{organization={Tianjin University},
	%                addressline={Street 29}, 
	%                postcode={1011 NX}, 
	%                postcodesep={}, 
	%                city={Amsterdam},
	country={China}}
\author[4]{Dongwei Ren}   %[orcid=0000-0002-0965-6810]
%\cormark[1]
%\credit{Writing -- review \& editing, Resources, Supervision}

\cortext[cor1]{Corresponding author}
%% Abstract
\begin{abstract}
Efficient single-image super-resolution (SISR) can benefit from allocating computation according to spatially varying reconstruction difficulty, rather than uniformly processing the entire image. In this paper, we propose a frequency-guided sparse computation framework that derives input-adaptive execution policies without introducing learnable routing modules. The policy is constructed directly from the frequency responses of each input image through non-parametric response partitioning and morphological context propagation. Under the evaluated \revise{synthetic }degradation shift\revise{s}, this design yields more stable computation allocation than the compared \revise{adaptive} methods. The resulting binary mask activates computation around frequency-responsive structures while bypassing redundant regions. We further \revise{decouple} the shared non-parametric policy construction \revise{from} architecture-specific sparse execution mechanisms for convolutional and Transformer-based SR backbones. For CNNs, dense convolution is reformulated as a masked projection over active local neighborhoods. For window-based Transformers, the pixel-level policy is aggregated into window-level execution decisions, allowing inactive windows to bypass the attention and feed-forward branches during inference. Across five CNN and Transformer backbones, the proposed implementations reduce floating-point operations (FLOPs) by 23.6–49.2\% and measured runtime by \revise{10.6–32.0\%}, while maintaining comparable reconstruction quality.

\end{abstract}


%	\item Self-supervised SelfDRSC++ corrects RS using dual reversed RS images, no GT GS videos.
%	\item RS reconstruction formulated as video frame interpolation.
%	\item Enables one-stage self-supervised training and reduces boundary artifacts.
%	\item Compact lightweight RS correction network surpasses SOTA on real-world textures.

%% Keywords
\begin{keyword}
Single-image super-resolution \sep Content-adaptive inference \sep Sparse computation \sep Frequency prior
\end{keyword}

\end{frontmatter}



\section{Introduction}

Single-image super-resolution (SISR) aims to reconstruct a high-resolution (HR) image from a low-resolution (LR) observation.
Modern CNN- and Transformer-based SR models have achieved strong reconstruction performance by increasing network depth and representational capacity~\cite{ledig2017photo,chen2023activating}.
However, their computational cost grows rapidly with spatial resolution, limiting their deployment in resource-constrained and latency-sensitive applications.

Most SR networks allocate the same amount of computation to every spatial location, despite substantial spatial variation in image content.
Textures and sharp structures exhibit pronounced frequency responses, whereas relatively smooth regions often occupy a considerable portion of an image.
Previous studies have therefore explored frequency or edge cues as indicators for spatially adaptive processing~\cite{wang2021exploring,xie2021learning}.
This observation suggests that uniform feature transformation may not be necessary at all locations and motivates input-adaptive computation that preserves sufficient processing for informative structures while reducing redundant operations elsewhere.

\begin{figure*}[!htbp]\footnotesize
	\centering
	\includegraphics[width=\linewidth]{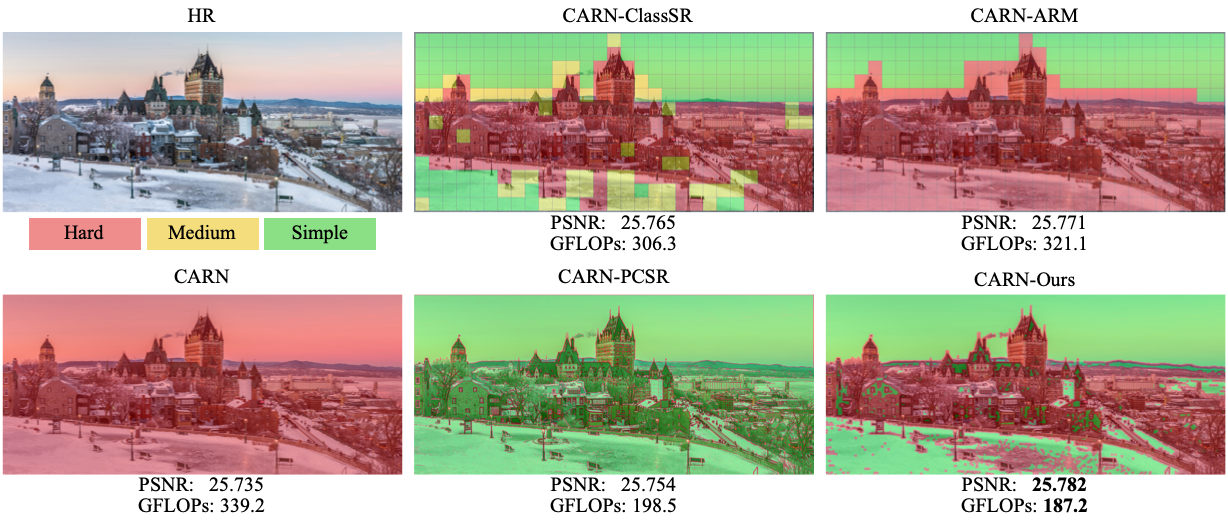}   
	\caption{
		Visualization of $\times 4$ SR results and spatial computation allocation for image ``1400'' from Test4K.
		The colored overlays visualize the computation assigned by different adaptive SR methods, where the colors indicate relative computation levels within each method.
		The proposed policy retains full feature processing around frequency-responsive structures while bypassing a large fraction of smoother regions, thereby achieving comparable reconstruction quality with lower computational cost.
	}
	\label{fig:intro}
\end{figure*}

Existing adaptive SR methods implement spatially varying computation through patch-level classification, lookup-based routing, early exiting, or pixel-level prediction~\cite{kong2021classsr,chen2022arm,wang2022adaptive,jeong2025accelerating}.
Patch-level approaches assign different networks or sub-networks to image patches, but their relatively coarse spatial granularity can limit fine-grained computation allocation.
Pixel-level routing provides more flexible decisions but commonly relies on an auxiliary predictor that must be trained jointly with or separately from the SR network.
Moreover, preprocessing, patch dispatch, and routing overhead may prevent reductions in theoretical FLOPs from translating into proportional runtime improvements.
These limitations motivate a lightweight policy that is input-dependent, spatially fine-grained, and directly compatible with efficient execution.
As illustrated in Fig.~\ref{fig:intro}, the proposed policy allocates
computation around frequency-responsive structures while bypassing a
large fraction of smoother regions, thereby maintaining comparable
reconstruction quality with lower computational cost.

To this end, we propose a frequency-guided sparse-computation framework that separates \emph{where to compute} from \emph{how to execute}.
Instead of learning an auxiliary model to predict reconstruction difficulty, we construct the execution policy directly from the frequency statistics of each LR input.
Specifically, the policy generator first computes a Gaussian high-pass magnitude response and applies per-image two-cluster quantization to partition the normalized response distribution.
The higher-response cluster provides a set of content-dependent active seeds.
Because spatial feature transformation also relies on the neighborhood surrounding these seeds, we further expand their support through morphological context propagation.
The propagation range controls the quality--computation trade-off: a wider range retains more contextual locations, whereas a narrower range produces a sparser execution policy.
The policy generator is parameter-free and requires no auxiliary router training. The SR backbone is nevertheless trained with the proposed masked execution scheme. Once trained, its test-time operating point can be adjusted through $K$ and $\sigma$ without retraining.
%\begin{table}[!htbp] %\footnotesize
%	\centering
%	\caption{Characteristic comparison over prior works.}
%	\label{table:intro_compare}
%	\resizebox{\linewidth}{!}{
	%		\begin{tabular}{l|ccccc}
		%			\hline
		%			{Characteristics} &ClassSR  &ARM &APE & PCSR  & Ours \\
		%			\midrule
		%			{w/o learnable routing} & \ding{55}  &  \Checkmark  &  \ding{55}  & \ding{55}  & \Checkmark  \\
		%			\midrule
		%			{w/o post-processing}  &  \ding{55}  & \ding{55}   & \ding{55}  & \ding{55}   &  \Checkmark \\
		%			\midrule
		%			Pixel-distributing  &  \ding{55} &  \ding{55} & \ding{55}  &  \Checkmark  & \Checkmark   \\
		%			\midrule
		%			Single-stage training & \ding{55}  &  \Checkmark  &  \Checkmark & \ding{55}  & \Checkmark   \\
		%			\bottomrule
		%		\end{tabular}
	%	}
%\end{table}
The resulting policy is instantiated through two architecture-specific sparse-execution schemes.
For convolutional networks, the proposed implementation gathers local descriptors at active spatial locations, processes them through the convolutional branch, and scatters the resulting features back to their original positions, while inactive locations follow a lightweight bypass path.
For window-based Transformers, the pixel-level policy is aggregated into window-level decisions, allowing inactive windows to skip the attention and feed-forward branches during inference.
The proposed framework therefore shares the same input-derived policy across different backbone families while adapting its execution mechanism to their respective operators.

We evaluate the proposed framework using three convolutional backbones, including CARN, SRResNet, and RCAN, and two Transformer backbones, including SwinIR and HAT.
Experiments on Test2K, Test4K, and Test8K show that the proposed implementations reduce FLOPs by 23.6--49.2\% and measured runtime by \revise{10.6--32.0\%}, while maintaining comparable PSNR, SSIM, and LPIPS.
Under the evaluated \revise{degradation types and severity levels}, the input-derived policy also exhibits a smaller change in allocated computation than the \revise{compared adaptive} baselines, indicating a more stable quality--computation trade-off in the tested setting.

\revise{We additionally compare CAMixerSR-B and ENAF and report independent-seed reconstruction statistics, repeated runtime, peak GPU memory, CPU efficiency, and operator-level profiling. Dedicated failure cases characterize where the frequency prior can under- or over-allocate computation.}

Our contributions are summarized as follows:
\begin{itemize}
	\item We introduce a router-free policy construction that combines per-image frequency-response partitioning with morphological context propagation, enabling input-adaptive computation without training an auxiliary routing network. \revise{The contribution lies in the router-free policy framework and its integration with sparse execution, rather than the individual classical image-processing operations.}
	
	\item We decouple policy construction from sparse execution and instantiate the shared policy through active-site computation for convolutional networks and window-level conditional execution for Transformers.
	
	\item We evaluate the proposed framework across five SR backbones and multiple image resolutions, demonstrating favorable reconstruction quality, computation, and measured-runtime trade-offs, together with more stable computation allocation under the evaluated degradation \revise{types and severities}. \revise{Repeated-run, practical-efficiency, and failure-case analyses further characterize these trade-offs and their limitations.}
\end{itemize}

\section{Related Work}
\subsection{Efficient SISR with Static Computation}

Deep neural networks have substantially advanced single-image super-resolution.
Early CNN-based methods established end-to-end mappings from LR to HR images~\cite{dong2014learning}, while subsequent residual and densely connected architectures further improved reconstruction performance by increasing network depth and capacity~\cite{lim2017enhanced,ledig2017photo,zhang2018image}.
More recently, Transformer-based models have employed windowed self-attention and enlarged receptive fields to capture long-range image dependencies~\cite{chen2021pre,chen2023activating,long2025progressive}.
Despite their strong reconstruction capability, these models generally incur considerable computational cost, particularly for high-resolution inputs.

A major line of research improves efficiency by designing compact, input-independent SR models.
FSRCNN~\cite{dong2016accelerating} performs most feature transformation in the LR space, while CARN~\cite{ahn2018fast} employs cascading residual connections to balance reconstruction accuracy and model complexity.
Other approaches search for efficient architectures through neural architecture search~\cite{lee2020journey,zhan2021achieving}, reduce numerical precision through quantization~\cite{li2020pams,xin2020binarized,liu20242dquant}, or transfer knowledge from larger models through distillation~\cite{lee2020learning,zhang2021data}.
Frequency information has also been incorporated into efficient SR
models. Li~\etal~\cite{li2025exploring} introduced frequency-guided
post-training quantization for hybrid convolutional and Transformer SR
models. More recently, Ying~\etal~\cite{ying2026frequency} proposed a
frequency-driven lattice network that reorganizes channel-wise and
global representations according to frequency features under
low-computation constraints. These methods improve the efficiency of a
fixed computation graph through model compression or lightweight
feature transformation, but generally apply the same execution
structure to different input images and spatial locations.

\subsection{Input-Adaptive Computation for SISR}

Input-adaptive SR methods exploit the observation that image content and reconstruction demand vary spatially.
SMSR~\cite{wang2021exploring} learns spatial and channel sparse masks to avoid redundant feature computation, while FAD~\cite{xie2021learning} \revise{introduces learned frequency-aware masks and multiple computational branches into SR backbones} to assign different computational capacities to different image regions.
These studies demonstrate the potential of spatial frequency cues for efficient SR, but their execution policies are learned jointly with the reconstruction network.

Other methods allocate computation at the patch, layer, or pixel level.
ClassSR~\cite{kong2021classsr} trains a classifier to divide image patches into different difficulty classes and processes them using backbone networks with different capacities.
ARM~\cite{chen2022arm} constructs a supernet containing sub-networks of different sizes and selects them for individual patches according to an Edge-to-PSNR lookup table.
APE~\cite{wang2022adaptive} predicts the incremental reconstruction capacity of successive layers and terminates computation early for selected patches.
More recently, PCSR~\cite{jeong2025accelerating} introduces a pixel-level classifier after feature extraction and directs pixels to upsampling branches with different widths.
In contrast, our policy specifies the spatial support on which repeated transformations in the network body are executed while retaining a shared feature-extraction and reconstruction path.

Recent studies further explore content-adaptive processing from different perspectives. CAMixerSR~\cite{wang2024camixersr} employs a learnable predictor to assign convolution or additional deformable window attention according to local image content. ENAF~\cite{nguyen2025enaf} combines multiple early exits with a learned PSNR estimator and adaptive patch fusion for large-image SR. AdaBM~\cite{hong2024adabm} adapts image- and layer-wise bit widths, representing a complementary form of adaptive inference through numerical precision rather than spatial execution. From the perspective of representation allocation, Qiao~\etal~\cite{qiao2026hierarchical} proposed a hierarchical
allocation framework that organizes lightweight feature enhancement
through branch-level pre-allocation and allocation-conditioned
region-level refinement. This method demonstrates the benefit of
assigning different enhancement capacities to heterogeneous spatial
cues, but performs learned soft modulation within a specifically
designed lightweight architecture. In contrast, our method constructs
a non-parametric binary execution policy and applies conditional
computation to different existing backbone families.

\revise{Supplementary Table~S1 compares adaptive mechanisms. Gaussian filtering, clustering, and dilation are established operations; our contribution is their integration into a shared router-free body-execution policy for CNN and Transformer backbones. FAD's learned masks and multiple branches alter the backbone and training objective, so FAD is compared conceptually rather than as an interchangeable module.}

\section{Proposed Method}
\label{sec:method}

\subsection{Overview and Design Principle}

Given a low-resolution input image
$\bm{x}\in\mathbb{R}^{3\times H\times W}$, we decompose an SR network
into a shallow feature extractor $\mathcal{H}$, a feature-extraction
body consisting of $L$ blocks, and a reconstruction module
$\mathcal{T}$:
\begin{equation}
	\bm{f}_0=\mathcal{H}(\bm{x}), \qquad
	\bm{f}_l=\mathcal{B}_l(\bm{f}_{l-1};\bm{m}), \quad
	\hat{\bm{y}}=\mathcal{T}(\bm{f}_L)+\mathcal{S}(\bm{x}),
\end{equation}
where $\bm{m}\in\{0,1\}^{H\times W}$ is an image-dependent execution
policy, and $\mathcal{S}$ denotes the backbone-specific global shortcut and
is omitted when the original backbone does not employ one. The shallow feature extractor and reconstruction module remain
dense, whereas the repeated transformations in the network body are
conditionally evaluated according to $\bm{m}$.

Our design separates \emph{where to compute} from \emph{how to
	execute}. The former is determined by a lightweight pixel policy
derived directly from image-frequency responses. The latter is
realized through backbone-specific execution mechanisms: pixel-level
active-site projection for CNNs and window-level conditional execution
for Transformers. 
The pixel-level policy $\bm{m}$ does not require an auxiliary trainable
classifier and is generated only once for each image, after which it is
reused throughout the network body. For Transformer blocks, the
corresponding window-level policy $\bm{m}_{\mathrm{win}}$ is derived
from $\bm{m}$ according to the actual regular or shifted window
partition of each block. Consequently, the frequency-based policy
generation cost does not increase with the depth of the SR backbone.

During training, the frequency-response seed mask is computed solely from each synthetic LR image. The same spatial augmentation is applied to the LR patch and its seed mask. Morphological dilation with the scheduled propagation range $K$ is then performed during training. During
inference, the policy is generated directly from the observed LR input,
without requiring an HR reference or a learned routing module.

\begin{figure*}[!htbp]\footnotesize
	\centering
	\includegraphics[width=\textwidth]{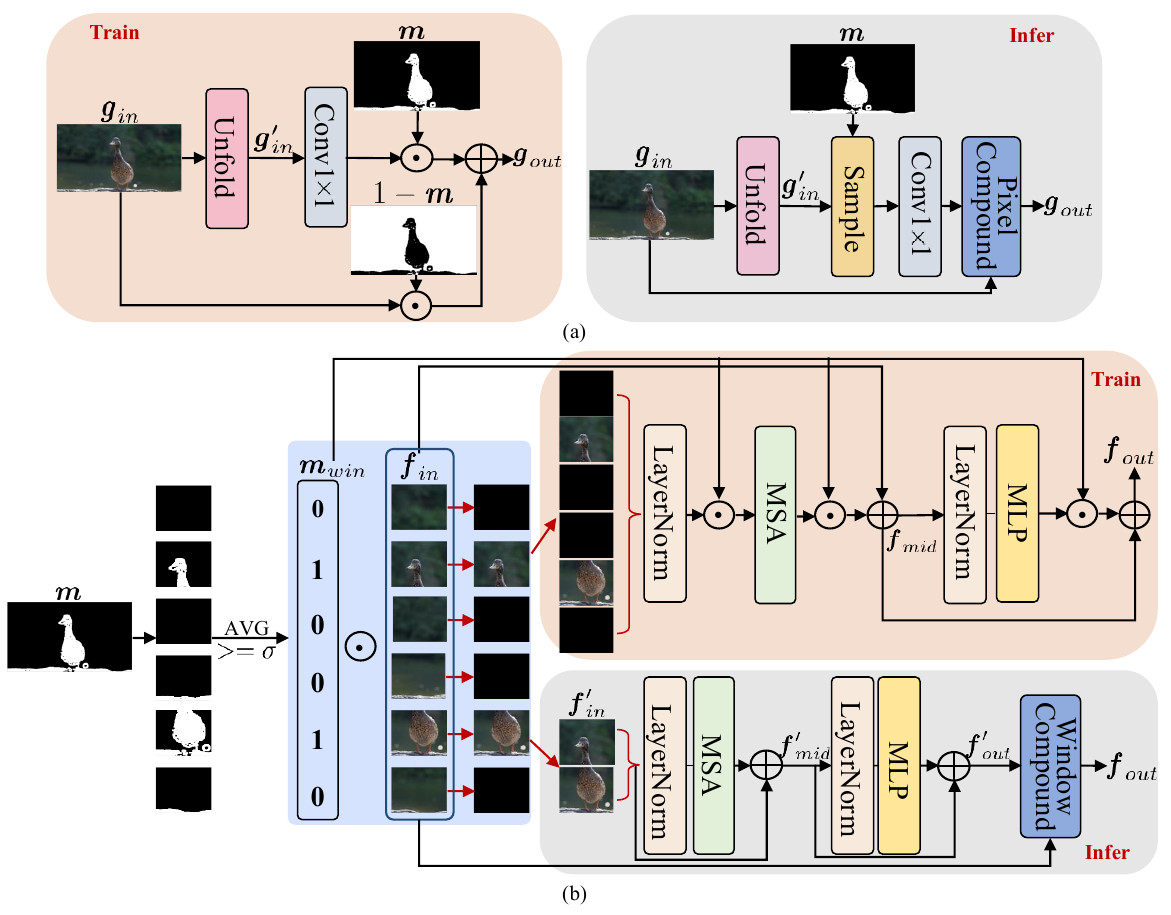}   
	\caption{\revise{Body-level adaptive execution during training and inference. (a) CNNs unfold complete neighborhoods, gate dense projection outputs during training, and gather/project/scatter active descriptors during inference. (b) Transformers aggregate pixel policies into window decisions. Training masks attention inputs after LayerNorm and gates attention/MLP outputs before residual addition. Inference processes active windows and merges them with inactive inputs. Identity paths remain unmasked.}}
	\label{fig:arch}
\end{figure*}
\subsection{Frequency-Guided Pixel Policy}
\label{sec:policy}

We first convert the input image into a grayscale image
$\bm{x}^{g}$ and compute its Gaussian high-pass magnitude response:
\begin{equation}
	\bm{h} = |\bm{x}^{g} - \mathcal{G}_{5,1} * \bm{x}^{g}|,
\end{equation}
where $\mathcal{G}_{5,1}$ denotes a $5\times5$ Gaussian kernel with
standard deviation $1$, and $*$ denotes spatial convolution. Accordingly, $\bm{h}$ is a non-negative Gaussian high-pass magnitude
response, where larger values indicate stronger local frequency
responses.

Because the response range varies considerably across images, we
normalize the response independently for each image:
\begin{equation}
	r_p
	=
	\frac{
		h_p-\min_q h_q
	}{
		\max_q h_q-\min_q h_q+\epsilon
	},
\end{equation}
where $p$ indexes a spatial location and $\epsilon$ is a small constant
that prevents numerical instability for nearly constant inputs.

\revise{Min--max normalization is a positive affine transformation, $r_p=a h_p+b$ ($a>0$), which scales the two-cluster objective by $a^2$ and preserves its optimal assignments apart from numerical precision. It serves numerical conditioning and comparable threshold baselines. Scores are relative within each image, not globally calibrated magnitudes.}

Applying a fixed threshold to $\bm{r}$ would impose the same decision
boundary on images with different response distributions. Instead, we
adaptively partition the scalar responses of each image into two
clusters:
\begin{equation}
	\left(
	\hat{\bm{z}},
	\hat{\mu}_0,
	\hat{\mu}_1
	\right)
	=
	\arg\min_{\bm{z},\mu_0,\mu_1}
	\sum_p
	\left(
	r_p-\mu_{z_p}
	\right)^2,
	\qquad
	z_p\in\{0,1\}.
\end{equation}
This objective performs per-image two-cluster quantization of the
response distribution. Unlike a learned classifier, its centroids are
estimated solely from the current image and are neither learned from
the training set nor predicted from intermediate network features.

The numerical labels returned by K-means have no inherent ordering.
We therefore identify the high-response cluster according to its
converged centroid:
\begin{equation}
	c_h
	=
	\arg\max_{c\in\{0,1\}}\hat{\mu}_c,
	\qquad
	s_p
	=
	\mathbb{I}
	\left[
	\hat{z}_p=c_h
	\right],
\end{equation}
where $\bm{s}$ denotes the initial frequency-derived seed mask.
Accordingly, the labels are explicitly remapped after clustering such
that $s_p=1$ always represents the cluster with the larger response
centroid. This definition is invariant to the arbitrary label ordering
of the K-means implementation.

The seed mask identifies locally responsive pixels but may omit the
surrounding context required by spatial feature transformations. We
therefore expand its support using binary morphological dilation:
\begin{equation}
	m^{(K)}_p
	=
	\max_{q\in\mathcal{B}_{K}(p)}
	s_q,
\end{equation}
where $\mathcal{B}_{K}(p)$ denotes the spatial support covered by a
$K\times K$ square structuring element anchored at location $p$,
using the fixed anchor
$(\lfloor K/2\rfloor,\lfloor K/2\rfloor)$ in our implementation.
Locations outside the image domain are omitted.
The resulting mask
$\bm{m}^{(K)}$ retains a controllable spatial margin around the
frequency-derived seeds. A larger $K$ includes more contextual
locations, whereas a smaller $K$ produces a more selective execution
policy. We use $\bm{m}=\bm{m}^{(K)}$ in the following sections.

During both training and inference, the policy is generated from the
LR input using the response partitioning and context-expansion
procedure described above. The policy is computed independently for
each image and requires no learned routing module.
% ------------------------------------------------

\subsection{Body-Level Active-Site Execution for CNNs}
\label{sec:cnn_sparse}
For CNN-based backbones, the pixel policy $\bm{m}$ determines the
spatial locations at which the computationally intensive body
transformations are evaluated. As illustrated in
Fig.~\ref{fig:arch}(a), we reformulate a $k\times k$ convolution as an
unfold operation followed by a shared $1\times1$ projection, thereby
supporting active-site computation without changing its receptive
field or learnable parameter count.

Given an input feature map
$\bm{g}_{\mathrm{in}}
\in
\mathbb{R}^{C_{\mathrm{in}}\times H\times W}$,
we first extract the local descriptors as
\begin{equation}
	\bm{g}'_{\mathrm{in}}
	=
	\mathtt{Unfold}_{k}
	\left(
	\bm{g}_{\mathrm{in}}
	\right)
	\in
	\mathbb{R}^{k^{2}C_{\mathrm{in}}\times H\times W},
	\label{eq:cnn_unfold}
\end{equation}
where $\mathtt{Unfold}_{k}(\cdot)$ collects the features within a
$k\times k$ neighborhood at every spatial location. A $k\times k$
convolution can subsequently be implemented by applying a
$1\times1$ convolution to $\bm{g}'_{\mathrm{in}}$:
\begin{equation}
	\mathcal{P}
	\left(
	\bm{g}'_{\mathrm{in}}
	\right)
	=
	\mathtt{Conv}_{1\times1}
	\left(
	\bm{g}'_{\mathrm{in}}
	\right).
	\label{eq:cnn_projection}
\end{equation}
The input channel dimension of the $1\times1$ projection is enlarged
by a factor of $k^{2}$, making this unfold--projection formulation
algebraically equivalent to the original $k\times k$ convolution.
When $k=1$, the unfold operation is unnecessary.

\revise{Our standard PyTorch implementation explicitly materializes $\mathtt{Unfold}$ before gathering active descriptors, without fused or custom CUDA kernels. This temporary tensor can exceed the input feature size; projection FLOPs savings therefore need not imply proportional runtime or memory savings. Section~\ref{sec:practical_efficiency} measures the complete inference path.}

For a same-dimensional body transformation, the training-time
conditional mapping is formulated as
\begin{equation}
	\bm{g}_{\mathrm{out}}
	=
	\bm{m}
	\odot
	\mathcal{P}
	\left(
	\bm{g}'_{\mathrm{in}}
	\right)
	+
	\left(
	\bm{1}-\bm{m}
	\right)
	\odot
	\bm{g}_{\mathrm{in}},
	\label{eq:cnn_train}
\end{equation}
where $\bm{m}$ is broadcast along the channel dimension. Active
locations receive the projected features, whereas inactive locations
retain their input features through the identity path. The
multiplication nodes in Fig.~\ref{fig:arch}(a) schematically represent
the two mask-controlled branches in Eq.~\eqref{eq:cnn_train}. Importantly, the pixel policy $\bm{m}$ selects output locations rather than masking the input features within each local neighborhood. For every active location $p$, the projection operates on the complete $k\times k$ descriptor extracted from $\bm{g}_{\mathrm{in}}$, including features from neighboring locations whose policy values may be zero.

During training, the unfold and projection operations retain dense
tensor shapes to support efficient mini-batch optimization. The
projection output is gated by $\bm{m}$, while the discrete policy is
not optimized through backpropagation. Gradients are used only to
update the SR backbone.

During inference, we avoid evaluating the projection at inactive
locations. Let
\begin{equation}
	\mathcal{A}
	=
	\left\{
	p\mid m_p=1
	\right\},
	\qquad
	Q=|\mathcal{A}|,
	\label{eq:active_set}
\end{equation}
where $\mathcal{A}$ denotes the active spatial locations. We sample
the corresponding descriptors from the unfolded feature:
\begin{equation}
	\bm{g}'_{\mathrm{in},\mathcal{A}}
	=
	\mathtt{Sample}
	\left(
	\bm{g}'_{\mathrm{in}},
	\bm{m}
	\right).
	\label{eq:cnn_sample}
\end{equation}
Only the sampled descriptors are processed by the $1\times1$
projection:
\begin{equation}
	\bm{g}'_{\mathrm{out},\mathcal{A}}
	=
	\mathtt{Conv}_{1\times1}
	\left(
	\bm{g}'_{\mathrm{in},\mathcal{A}}
	\right).
	\label{eq:cnn_active_projection}
\end{equation}
Finally, the processed features are placed back at their original
locations, while inactive locations preserve
$\bm{g}_{\mathrm{in}}$:
\begin{equation}
	\bm{g}_{\mathrm{out}}
	=
	\mathtt{PixelCompound}
	\left(
	\bm{g}'_{\mathrm{out},\mathcal{A}},
	\bm{g}_{\mathrm{in}},
	\bm{m}
	\right).
	\label{eq:cnn_compound}
\end{equation}
Here, $\mathtt{PixelCompound}(\cdot)$ denotes the scatter-and-merge
operation shown in Fig.~\ref{fig:arch}(a). 
Since the sampling operation selects complete descriptor columns after neighborhood extraction, Eqs. \eqref{eq:cnn_sample}–\eqref{eq:cnn_compound} exactly reproduce the training-time mapping in Eq. \eqref{eq:cnn_train}, while avoiding projection at inactive output locations.

\begin{sloppypar}
The dominant multiply--accumulate cost of the projection is reduced
from $HWk^{2}C_{\mathrm{in}}C_{\mathrm{out}}$ to $Qk^{2}C_{\mathrm{in}}C_{\mathrm{out}}$.
Its computation ratio is therefore approximately $Q/(HW)$.
The practical costs of unfolding, active-site sampling, coordinate
generation, and feature composition are included in the measured
runtime. The shallow feature extractor, channel-changing layers,
lightweight feature-fusion operations, and reconstruction module
remain dense according to the original backbone design.
\end{sloppypar}

% ------------------------------------------------
\subsection{Window-Level Conditional Execution for Transformers}

Window-based Transformers jointly process the feature tokens within
each attention window. We therefore use a complete feature window as
the basic execution unit, rather than pruning individual tokens within
a window. As illustrated in Fig.~\ref{fig:arch}(b), the pixel-level
policy $\bm{m}$ is first partitioned according to the window layout and
then aggregated into a binary window-level policy
$\bm{m}_{\mathrm{win}}$.

Let $\mathcal{W}_{j}$ denote the spatial support of the $j$-th window.
The proportion of active pixels within this window is computed as
\begin{equation}
	\bar{m}_{j}
	=
	\frac{1}{|\mathcal{W}_{j}|}
	\sum_{p\in\mathcal{W}_{j}}m_{p}.
\end{equation}
The corresponding window-level execution decision is defined as
\begin{equation}
	m_{\mathrm{win},j}
	=
	\mathbb{I}
	\left[
	\bar{m}_{j}\geq\sigma
	\right],
\end{equation}
where $\sigma$ denotes the window activation threshold. Collecting the
decisions of all windows gives
\begin{equation}
	\bm{m}_{\mathrm{win}}
	=
	\left[
	m_{\mathrm{win},1},
	m_{\mathrm{win},2},
	\ldots,
	m_{\mathrm{win},N_{\mathrm{win}}}
	\right]^{\mathsf T}.
\end{equation}
A value of $m_{\mathrm{win},j}=1$ identifies an active window, whereas
$m_{\mathrm{win},j}=0$ identifies a window that can be omitted during
inference. For regular and shifted-window blocks, the pixel policy is
partitioned according to the actual window layout of the corresponding
block. For clarity, the block indices are omitted in the following
equations.

Let $\bm{f}_{\mathrm{in}}$ denote the input feature windows of a
Transformer block. During training, all windows are retained in a dense
tensor layout to support efficient mini-batch optimization.
\revise{For this representative attention and feed-forward residual block, the attention input is masked after LayerNorm, and the attention output is explicitly gated before residual addition:}
\begin{equation}
    \bm{f}_{\mathrm{mid}}
    =\revise{\bm{m}_{\mathrm{win}}\odot}
    \operatorname{MSA}\!\left(
    \revise{\operatorname{LN}(\bm{f}_{\mathrm{in}})\odot\bm{m}_{\mathrm{win}}}
    \right)+\bm{f}_{\mathrm{in}}.
    \label{eq:transformer_attention_gate}
\end{equation}
\revise{The feed-forward output is similarly gated:}
\begin{equation}
    \bm{f}_{\mathrm{out}}
    =\revise{\bm{m}_{\mathrm{win}}\odot}
    \operatorname{MLP}\!\left(\operatorname{LN}(\bm{f}_{\mathrm{mid}})\right)
    +\bm{f}_{\mathrm{mid}}.
    \label{eq:transformer_mlp_gate}
\end{equation}
Here, each element of $\bm{m}_{\mathrm{win}}$ is broadcast to all
tokens and channels within its corresponding window.
\revise{LayerNorm affine parameters and projection biases can produce nonzero internal outputs even for zero inputs, so input masking alone does not guarantee an inactive branch. Output gating instead gives, for $m_{\mathrm{win},j}=0$,}
\revise{
\begin{equation}
    \bm{f}_{\mathrm{out},j}
    =\bm{f}_{\mathrm{mid},j}
    =\bm{f}_{\mathrm{in},j}.
    \label{eq:transformer_inactive_identity}
\end{equation}}
\revise{An inactive window therefore follows an exact identity mapping regardless of the internal LayerNorm, MSA, or MLP outputs.}
\revise{For active windows, both equations reduce to the standard attention and feed-forward transformations.}
The residual identity paths remain unmasked in both transformations.

\revise{This output gating is used for SwinIR and HAT. HAT's convolutional and channel-attention operations use CNN-style conditional execution, and its window-based branches follow the formulation above.}

During inference, the inactive windows are directly omitted from the
attention and feed-forward branches. We first sample the active feature
windows and organize them into a compact batch:
\begin{equation}
	\bm{f}'_{\mathrm{in}}
	=
	\operatorname{Sample}
	\left(
	\bm{f}_{\mathrm{in}},
	\bm{m}_{\mathrm{win}}
	\right).
\end{equation}
The sampled windows are then processed by the standard Transformer
transformations:
\begin{equation}
	\bm{f}'_{\mathrm{mid}}
	=
	\operatorname{MSA}
	\left(
	\operatorname{LN}
	\left(
	\bm{f}'_{\mathrm{in}}
	\right)
	\right)
	+
	\bm{f}'_{\mathrm{in}},
\end{equation}
\begin{equation}
	\bm{f}'_{\mathrm{out}}
	=
	\operatorname{MLP}
	\left(
	\operatorname{LN}
	\left(
	\bm{f}'_{\mathrm{mid}}
	\right)
	\right)
	+
	\bm{f}'_{\mathrm{mid}}.
\end{equation}
Finally, the processed active windows are placed back at their original
locations, while inactive windows retain the corresponding input
features from $\bm{f}_{\mathrm{in}}$:
\begin{equation}
	\bm{f}_{\mathrm{out}}
	=
	\operatorname{WindowCompound}
	\left(
	\bm{f}'_{\mathrm{out}},
	\bm{f}_{\mathrm{in}},
	\bm{m}_{\mathrm{win}}
	\right).
\end{equation}
Here, $\operatorname{WindowCompound}(\cdot)$ denotes the window-level
scatter-and-merge operation illustrated in Fig.~\ref{fig:arch}(b).
Accordingly, training retains a regular dense tensor layout with
\revise{policy-gated branch outputs}, whereas inference realizes actual
computation reduction by executing the attention and feed-forward
branches only for active windows.

\revise{Inactive windows thus retain their input features exactly in both the masked training-time mapping and sparse inference.}

\subsection{Test-Time Computation Adjustment}
\label{sec:computation_adjustment}

Both $K$ and $\sigma$ can be adjusted during inference without
retraining the SR network or updating its learned weights. They
therefore provide an explicit test-time trade-off between reconstruction
quality and computational cost. Since the resulting computation also
depends on the image content and backbone architecture, we report both
FLOPs and measured runtime in the experiments.

% ------------------------------------------------

\section{Experiments}

\subsection{Experimental Settings}
%------------------------------------------------------------------------
\subsubsection{Datasets}
\label{sec:dataset}
All models are trained on DIV2K~\cite{agustsson2017ntire}, which comprises 800 high-resolution images. Following the training setting in~\cite{kong2021classsr,chen2022arm, jeong2025accelerating}, we partition these images into 1.59 million $32\times 32$ LR sub-images for training. Random rotation and flipping are applied for data augmentation.

\revise{DIV2K validation images 0801--0900 select inference operating points; they are disjoint from training and the DIV8K-derived test benchmarks, which are used only for final evaluation.}
For the testing dataset, we select 300 images (index $1201-1500$) from DIV8K~\cite{gu2019div8k}. Following~\cite{kong2021classsr,chen2022arm, jeong2025accelerating}, we divide these three hundred images evenly into three groups. The first two groups are downsampled to 2K and 4K resolution, respectively, to serve as HR images of Test2K and Test4K. The third group is used directly as HR images of Test8K.
The LR images are obtained following the previous work~\cite{kong2021classsr}. 
To verify that the proposed policy remains effective at low
resolution, we also conduct experiments on
Set5~\cite{bevilacqua2012low}, Set14~\cite{zeyde2010single}, and
Urban100~\cite{huang2015single}.
%	

%-------------------------------------------------------------------------
\subsubsection{Implementation Details}
\label{sec:implementation_details}
Following existing adaptive SR methods~\cite{kong2021classsr,chen2022arm,jeong2025accelerating}, we adopt the same CARN~\cite{ahn2018fast} and SRResNet~\cite{ledig2017photo} configurations for direct and fair comparison. We further evaluate RCAN~\cite{zhang2018image}, SwinIR~\cite{liang2021swinir}, and HAT~\cite{chen2023activating} to examine the generality of the proposed framework across CNN- and Transformer-based backbones, following their standard implementations.

The training batch size is set to 16. The initial learning rate is set as $2\times 10^{-4}$ with cosine annealing scheduling~\cite{loshchilov2016sgdr}. 
We adopt the $\mathcal{L}_1$ loss and the Adam optimizer~\cite{kingma2014adam}, and train all models for 2 million iterations. Runtime is measured with an inference batch size of 1 on a single
NVIDIA RTX A6000 GPU \revise{(48 GB)}. \revise{We use PyTorch 2.11.0+cu128 and CUDA 12.8. After one warm-up, 10 runs with CUDA synchronization give mean $\pm$ standard deviation.} All methods are evaluated under the same
single-stream inference setting. For patch-based \revise{or tile-based} baselines, the
reported runtime includes patch partitioning, routing \revise{or exit selection}, branch-wise
processing, and output stitching. Routed patch groups are processed
sequentially, and no method-specific cross-branch or multi-stream
parallelization is introduced for any method.

During training, the propagation range is progressively reduced within
a backbone-specific range to transition from spatially smooth policies
to more localized and selective computation. Specifically, $K$ is
reduced from 5 to 2 for CARN, from 11 to 5 for SRResNet and RCAN, and
from 11 to 3 for SwinIR and HAT. The transition schedule is fixed for each backbone and kept unchanged throughout all corresponding experiments. \revise{These empirical curriculum ranges are fixed before test evaluation; fixed- and random-$K$ controls are in Section~\ref{sec:training_schedule}.}
\revise{On DIV2K validation, candidate $K$ values (CNNs) and $(K,\sigma)$ pairs (Transformers) are compared by quality--computation trade-off, selecting settings with quality close to full computation and substantial savings.} We set $K=5$ for CARN,
$K=7$ for SRResNet, and $K=11$ for RCAN, SwinIR, and HAT. The window
activation threshold is set to $\sigma=0.5$ for Transformer-based
models. Each adaptive model is trained using the proposed masked
computation scheme, while different test-time values of $K$ and
$\sigma$ are evaluated using the same trained checkpoint. Therefore,
the quality--computation trade-off can be adjusted at inference without
retraining the network. \revise{The selected settings are fixed for Test2K, Test4K, and Test8K without further adjustment. All selected $K$ values lie within the overall ranges encountered during training.}

\subsubsection{Evaluation Metrics}
We evaluate reconstruction quality using PSNR, SSIM, and LPIPS. PSNR and SSIM measure pixel-wise fidelity and structural similarity between the reconstructed SR images and their HR references, respectively. LPIPS is additionally used to assess perceptual similarity. For efficiency evaluation, we report FLOPs and runtime. A favorable acceleration method should reduce computational cost and runtime while maintaining competitive distortion and perceptual quality.

\revise{We additionally report peak allocated GPU memory, component-level runtime, and single-core CPU inference. Five independent training seeds are evaluated for the SRResNet-based reconstruction comparison, and GPU runtime includes repeated-run statistics.}

\begin{table}[!htbp]
\footnotesize
\setlength{\tabcolsep}{3pt}
\centering
\caption{\revise{Comparison on Test2K, Test4K, and Test8K for $\times4$ SR. ENAF is evaluated with the CNN backbone families, while CAMixerSR-B is an independent architecture. Best and second-best results within each backbone group are in bold and underlined, respectively.}}
\label{table:sota}
\resizebox{\linewidth}{!}{
\begin{tabular}{l|c|ccc|ccc|ccc}
\hline
\hline
\multirow{2}{*}{Methods} & \multirow{2}{*}{Params}
& \multicolumn{3}{c|}{Test2K}
& \multicolumn{3}{c|}{Test4K}
& \multicolumn{3}{c}{Test8K} \\
\cline{3-11}
& & PSNR & SSIM & LPIPS
& PSNR & SSIM & LPIPS
& PSNR & SSIM & LPIPS \\
\hline

CARN
& 295K
& 26.022 & 0.7523 & 0.2849
& 27.405 & 0.7988 & 0.2604
& 33.141 & 0.8745 & \underline{0.2007} \\

CARN-ClassSR
& 645K
& 26.020 & \underline{0.7543} & \underline{0.2848}
& \underline{27.430} & \underline{0.7996} & 0.2606
& 33.252 & \underline{0.8749} & 0.2008 \\

CARN-ARM
& 295K
& \underline{26.029} & 0.7540 & 0.2850
& \underline{27.430} & 0.7992 & \underline{0.2602}
& 33.279 & \underline{0.8749} & 0.2010 \\

CARN-PCSR
& 321K
& 26.023 & 0.7519 & 0.2853
& 27.429 & 0.7982 & 0.2609
& \underline{33.283} & 0.8745 & 0.2013 \\

CARN-ENAF
& 395K
& 26.028 & 0.7538 & 0.2850
& 27.429 & 0.7990 & 0.2604
& 33.280 & 0.8748 & 0.2010 \\

CARN-Ours
& 295K
& \textbf{26.034} & \textbf{0.7544} & \textbf{0.2845}
& \textbf{27.433} & \textbf{0.7998} & \textbf{0.2600}
& \textbf{33.284} & \textbf{0.8750} & \textbf{0.2003} \\

\hline

SRResNet
& 1.5M
& \textbf{26.223} & 0.7609 & 0.2633
& 27.657 & \underline{0.8067} & \underline{0.2378}
& 33.416 & \textbf{0.8793} & \underline{0.1823} \\

SRResNet-ClassSR
& 3.1M
& 26.212 & \textbf{0.7623} & \underline{0.2631}
& 27.663 & \textbf{0.8068} & 0.2379
& 33.508 & \textbf{0.8793} & 0.1824 \\

SRResNet-ARM
& 1.5M
& 26.215 & \underline{0.7620} & 0.2634
& \underline{27.669} & \textbf{0.8068} & 0.2382
& 33.520 & 0.8791 & 0.1828 \\

SRResNet-PCSR
& 1.5M
& \underline{26.216} & 0.7619 & 0.2637
& 27.662 & \underline{0.8067} & 0.2385
& \underline{33.528} & \underline{0.8792} & 0.1831 \\

SRResNet-ENAF
& 2.0M
& 26.215 & 0.7619 & 0.2633
& 27.668 & \underline{0.8067} & 0.2383
& 33.521 & 0.8791 & 0.1829 \\

SRResNet-Ours
& 1.5M
& \textbf{26.223} & \textbf{0.7623} & \textbf{0.2630}
& \textbf{27.670} & \textbf{0.8068} & \textbf{0.2375}
& \textbf{33.532} & \textbf{0.8793} & \textbf{0.1820} \\

\hline

RCAN
& 15.6M
& 26.393 & 0.7688 & \underline{0.2346}
& 27.880 & 0.8130 & \underline{0.2072}
& 33.723 & 0.8828 & \underline{0.1633} \\

RCAN-ClassSR
& 30.1M
& 26.396 & 0.7689 & 0.2350
& 27.879 & 0.8132 & 0.2075
& 33.722 & 0.8824 & 0.1635 \\

RCAN-ARM
& 15.6M
& \underline{26.398} & 0.7691 & 0.2347
& 27.880 & 0.8130 & 0.2075
& 33.725 & \underline{0.8833} & 0.1638 \\

RCAN-PCSR
& 15.6M
& \underline{26.398} & \underline{0.7696} & \underline{0.2346}
& \underline{27.884} & \underline{0.8134} & 0.2077
& \underline{33.728} & 0.8832 & 0.1640 \\

RCAN-ENAF
& 16.1M
& 26.397 & 0.7691 & 0.2348
& 27.881 & 0.8130 & 0.2075
& 33.724 & 0.8832 & 0.1639 \\

RCAN-Ours
& 15.6M
& \textbf{26.400} & \textbf{0.7698} & \textbf{0.2344}
& \textbf{27.886} & \textbf{0.8136} & \textbf{0.2070}
& \textbf{33.730} & \textbf{0.8835} & \textbf{0.1630} \\

\hline

SwinIR
& 930K
& \underline{26.323} & \textbf{0.7676} & \underline{0.2356}
& \underline{27.802} & \underline{0.8113} & \underline{0.2127}
& \underline{33.651} & \underline{0.8819} & \underline{0.1629} \\

SwinIR-Ours
& 930K
& \textbf{26.347} & \underline{0.7675} & \textbf{0.2351}
& \textbf{27.811} & \textbf{0.8114} & \textbf{0.2123}
& \textbf{33.666} & \textbf{0.8820} & \textbf{0.1626} \\

\hline

CAMixerSR-B
& 765K
& 26.336 & 0.7673 & 0.2358
& 27.809 & 0.8111 & 0.2129
& 33.663 & 0.8817 & 0.1630 \\

\hline

HAT
& 9.6M
& \underline{26.619} & \underline{0.7752} & \underline{0.2096}
& \underline{28.113} & \underline{0.8178} & \underline{0.1889}
& \underline{33.870} & \underline{0.8842} & \underline{0.1449} \\

HAT-Ours
& 9.6M
& \textbf{26.628} & \textbf{0.7754} & \textbf{0.2087}
& \textbf{28.120} & \textbf{0.8180} & \textbf{0.1881}
& \textbf{33.888} & \textbf{0.8845} & \textbf{0.1441} \\

\hline
\hline
\end{tabular}
}
\end{table}
\begin{table}[!htbp]
\footnotesize
\setlength{\tabcolsep}{3pt}
\centering
\setlength{\abovecaptionskip}{0pt}
\setlength{\belowcaptionskip}{2pt}
\caption{\revise{GPU runtime and FLOPs per image for $\times4$ SR. Each cell lists runtime (mean $\pm$ standard deviation over 10 runs on an RTX A6000) followed by FLOPs, separated by a slash. Percentages are relative to the dense backbone; CAMixerSR-B uses absolute FLOPs. Best runtime and FLOPs within each backbone group are bold.}}
\label{table:time}
\resizebox{\linewidth}{!}{%
\begin{tabular}{l|c|c|c}
\hline
\hline
Method & Test2K & Test4K & Test8K \\
\hline

CARN
& $91 \pm 3$ ms / 112.0G (100\%)
& $290 \pm 23$ ms / 457.8G (100\%)
& $1.62 \pm 0.10$ s / 2.639T (100\%) \\

CARN-ClassSR \cite{kong2021classsr}
& $699 \pm 33$ ms / 100.3G (89.6\%)
& $3609 \pm 102$ ms / 378.5G (82.7\%)
& $21.03 \pm 1.06$ s / 1.800T (68.2\%) \\

CARN-ARM \cite{chen2022arm}
& $691 \pm 30$ ms / 84.3G (75.3\%)
& $3150 \pm 98$ ms / 328.8G (71.8\%)
& $19.55 \pm 0.99$ s / 1.739T (65.9\%) \\

CARN-PCSR \cite{jeong2025accelerating}
& $90 \pm 5$ ms / 68.7G (61.3\%)
& $275 \pm 40$ ms / 277.7G (60.7\%)
& $1.40 \pm 0.17$ s / 1.646T (62.4\%) \\

CARN-ENAF \cite{nguyen2025enaf}
& $670 \pm 15$ ms / 75.3G (67.2\%)
& $2788 \pm 111$ ms / 312.9G (68.3\%)
& $16.50 \pm 1.03$ s / 1.668T (63.2\%) \\

CARN-Ours
& \textbf{\boldmath$76 \pm 2$ ms} / \textbf{64.7G (57.8\%)}
& \textbf{\boldmath$220 \pm 9$ ms} / \textbf{256.1G (55.9\%)}
& \textbf{\boldmath$1.23 \pm 0.12$ s} / \textbf{1.394T (52.8\%)} \\

\hline

SRResNet
& $119 \pm 6$ ms / 502.9G (100\%)
& $430 \pm 19$ ms / 2.056T (100\%)
& $2.50 \pm 0.23$ s / 11.851T (100\%) \\

SRResNet-ClassSR \cite{kong2021classsr}
& $550 \pm 22$ ms / 445.2G (88.5\%)
& $2733 \pm 104$ ms / 1.684T (81.9\%)
& $15.96 \pm 0.93$ s / 7.982T (67.4\%) \\

SRResNet-ARM \cite{chen2022arm}
& $533 \pm 30$ ms / 394.8G (78.5\%)
& $2403 \pm 114$ ms / 1.502T (73.1\%)
& $15.38 \pm 0.98$ s / 7.399T (62.4\%) \\

SRResNet-PCSR \cite{jeong2025accelerating}
& $154 \pm 8$ ms / 388.2G (77.2\%)
& $549 \pm 29$ ms / 1.487T (72.3\%)
& $3.00 \pm 0.23$ s / 7.352T (62.0\%) \\

SRResNet-ENAF \cite{nguyen2025enaf}
& $471 \pm 21$ ms / 390.1G (77.6\%)
& $2111 \pm 95$ ms / 1.498T (72.9\%)
& $14.02 \pm 0.94$ s / 7.358T (62.1\%) \\

SRResNet-Ours
& \textbf{\boldmath$98 \pm 7$ ms} / \textbf{349.6G (69.5\%)}
& \textbf{\boldmath$351 \pm 19$ ms} / \textbf{1.367T (66.5\%)}
& \textbf{\boldmath$1.70 \pm 0.20$ s} / \textbf{7.288T (61.5\%)} \\

\hline

RCAN
& $286 \pm 10$ ms / 3.153T (100\%)
& $1.41 \pm 0.08$ s / 12.078T (100\%)
& $11.5 \pm 0.77$ s / 69.623T (100\%) \\

RCAN-ClassSR \cite{kong2021classsr}
& $1005 \pm 30$ ms / 2.740T (86.9\%)
& $10.9 \pm 0.49$ s / 9.674T (80.1\%)
& $60.0 \pm 2.80$ s / 48.875T (70.2\%) \\

RCAN-ARM \cite{chen2022arm}
& $992 \pm 26$ ms / 2.598T (82.4\%)
& $9.59 \pm 0.37$ s / 9.397T (77.8\%)
& $59.1 \pm 2.1$ s / 46.020T (66.1\%) \\

RCAN-PCSR \cite{jeong2025accelerating}
& $244 \pm 14$ ms / 2.261T (71.7\%)
& $1.20 \pm 0.08$ s / 9.095T (75.3\%)
& $15.0 \pm 0.92$ s / 45.185T (64.9\%) \\

RCAN-ENAF \cite{nguyen2025enaf}
& $952 \pm 28$ ms / 2.507T (79.5\%)
& $9.30 \pm 0.35$ s / 9.299T (77.0\%)
& $57.3 \pm 2.0$ s / 46.641T (67.0\%) \\

RCAN-Ours
& \textbf{\boldmath$200 \pm 11$ ms} / \textbf{1.933T (61.3\%)}
& \textbf{\boldmath$1.01 \pm 0.12$ s} / \textbf{7.017T (58.1\%)}
& \textbf{\boldmath$8.0 \pm 0.83$ s} / \textbf{35.368T (50.8\%)} \\

\hline

SwinIR
& $358 \pm 34$ ms / 102.6G (100\%)
& $2.90 \pm 0.26$ s / 415.0G (100\%)
& $14.98 \pm 1.30$ s / 2.400T (100\%) \\

SwinIR-Ours
& \textbf{\boldmath$299 \pm 27$ ms} / \textbf{78.4G (76.4\%)}
& \textbf{\boldmath$2.20 \pm 0.20$ s} / \textbf{289.9G (69.9\%)}
& \textbf{\boldmath$10.82 \pm 0.96$ s} / \textbf{1.412T (58.8\%)} \\

\hline

CAMixerSR-B \cite{wang2024camixersr}
& $643 \pm 40$ ms / 80.2G
& $4.11 \pm 0.31$ s / 303.7G
& $19.55 \pm 1.56$ s / 1.608T \\

\hline

HAT
& $4.53 \pm 0.35$ s / 1.341T (100\%)
& $34.86 \pm 1.27$ s / 5.425T (100\%)
& $150.0 \pm 3.48$ s / 31.369T (100\%) \\

HAT-Ours
& \textbf{\boldmath$3.97 \pm 0.40$ s} / \textbf{1.008T (75.2\%)}
& \textbf{\boldmath$31.16 \pm 0.88$ s} / \textbf{3.749T (69.1\%)}
& \textbf{\boldmath$130.4 \pm 1.84$ s} / \textbf{17.943T (57.2\%)} \\

\hline
\hline
\end{tabular}
}
\end{table}

%-------------------------------------------------------------------------
\subsection{Comparison with State-of-the-art Methods}
Following experimental settings of ClassSR~\cite{kong2021classsr}, ARM~\cite{chen2022arm}, and PCSR~\cite{jeong2025accelerating}, we evaluate both performance and efficiency of our method on $\times 4$ SR, reporting the metrics of PSNR, SSIM, LPIPS, FLOPs, and runtime.
As shown by the quantitative results in
Table~\ref{table:sota}, our method maintains competitive reconstruction
quality while consistently reducing computational cost.
The attainable acceleration is closely related to the proportion of
total computation contributed by the conditionally executed network
body. Accordingly, the method provides a moderate FLOPs reduction for
SRResNet, whose reconstruction and upsampling modules account for a
larger share of FLOPs. Most FLOPs in RCAN and SwinIR are in the body.
RCAN saves more than SRResNet on all three benchmarks, and SwinIR only on Test8K. \revise{Under the GPU protocol in Table~\ref{table:time}, our method achieves lower measured runtime than the compared adaptive baselines. The single-core CPU results in Section~\ref{sec:practical_efficiency} show that this advantage is hardware- and implementation-dependent: SRResNet-Ours remains faster than dense SRResNet, but its CPU latency is comparable to ClassSR and ARM.}

\revise{ENAF~\cite{nguyen2025enaf} is backbone-matched for CARN and SRResNet; its multi-exit strategy is also applied to RCAN. CAMixerSR-B~\cite{wang2024camixersr} is an independent architecture. ENAF adds reconstruction exits and a learned quality estimator, increasing CARN/SRResNet parameters from 295K/1.5M to 395K/2.0M; our policy adds none.

Tables~\ref{table:sota} and~\ref{table:time} show comparable reconstruction quality with lower FLOPs and runtime than ENAF. On Test8K, CAMixerSR-B achieves 33.663 dB at 1.608T FLOPs and $19.55\pm1.56$ s, versus 33.666 dB, 1.412T, and $10.82\pm0.96$ s for SwinIR-Ours. This compares complete architectures rather than isolating their routing mechanisms. Our gathering and composition costs are examined in Section~\ref{sec:practical_efficiency}.}

\begin{figure}[!htbp]\footnotesize
	\centering
	\includegraphics[width=0.8\linewidth]{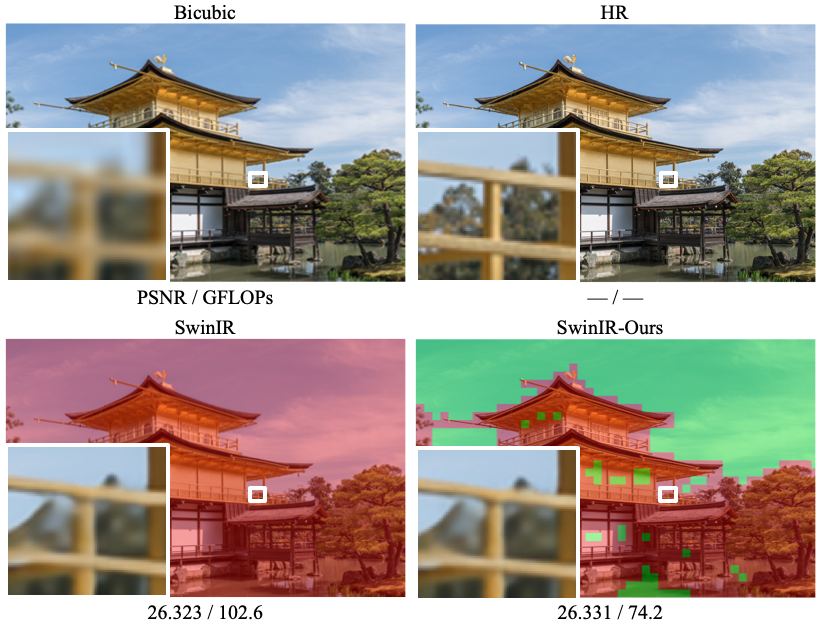}   %
	\caption{
		Visual comparison of our method with the backbone network. The colored overlays visualize the window-level computation policy, and the corresponding PSNR/GFLOPs are reported below each result.
	}
	\label{fig:swin}
\end{figure}

We also provide qualitative comparisons, together with the PSNR and FLOPs of each SR result, in \revise{Supplementary Fig.~S2} for more intuitive evaluation.
Furthermore, our method can be integrated into Transformer-based
architectures such as SwinIR with limited architecture-specific
modifications, achieving significant FLOPs reduction while maintaining comparable reconstruction quality, as shown in Fig.~\ref{fig:swin}. 
\subsubsection{\revise{Reconstruction Statistics and Training--Inference Controls}}

\revise{Supplementary Table~S2 reports five independent training seeds under the same protocol. The observed variation does not establish significant quality gains from the small mean PSNR differences; the results support comparable quality alongside reduced computation.}

\revise{Table~\ref{table:training_inference_control} compares conventional dense training/inference with all-active and sparse inference of the same sparse-trained SwinIR checkpoint. Differences between (i) and (ii) concern training and optimization, potentially including ordinary checkpoint variation. Comparing (ii) and (iii) isolates sparse inference: PSNR decreases by 0.009, 0.017, and 0.020 dB on Test2K, Test4K, and Test8K, respectively, with substantial FLOPs savings. The benefit is reduced computation with quality close to full execution.}

\begin{table}[!htbp]
\footnotesize
\setlength{\tabcolsep}{5pt}
\centering
\caption{\revise{SwinIR controls for training and inference. Settings (ii) and (iii) use the same sparse-trained checkpoint; only the inference execution policy differs. The all-active setting uses $\sigma=0$.}}
\label{table:training_inference_control}
\resizebox{\linewidth}{!}{
\begin{tabular}{l|cc|cc|cc}
\hline
\hline
\multirow{2}{*}{Setting}
& \multicolumn{2}{c|}{Test2K}
& \multicolumn{2}{c|}{Test4K}
& \multicolumn{2}{c}{Test8K} \\
& PSNR & GFLOPs
& PSNR & GFLOPs
& PSNR & GFLOPs \\
\hline

(i) Dense training + dense inference
& 26.323 & 102.6
& 27.802 & 415.0
& 33.651 & 2399.8 \\

(ii) Sparse training + all-active inference
& 26.356 & 102.6
& 27.828 & 415.0
& 33.686 & 2399.8 \\

(iii) Sparse training + proposed sparse inference
& 26.347 & 78.4
& 27.811 & 289.9
& 33.666 & 1412.0 \\

\hline
\hline
\end{tabular}
}
\end{table}

\subsection{\revise{Practical Efficiency}}
\label{sec:practical_efficiency}

\revise{Supplementary Tables~S3--S5 detail runtime components, peak GPU memory, and single-core CPU inference. On Test2K, policy generation takes 2.3\% of SRResNet-Ours runtime and 1.2\% of SwinIR-Ours runtime, while Unfold/gather and window policy/gather take 46.3\% and 80.5\%, respectively. These are absolute component shares, not incremental overhead over dense inference. This data-movement cost is consistent with the gap between FLOPs and runtime savings.

After warm-up and resetting CUDA peak statistics, complete-forward peak memory decreases from 2.02/8.05/32.16 GB to 1.86/7.43/29.70 GB for whole-image SRResNet on Test2K/Test4K/Test8K. This is a net measurement, not a per-layer attribution; materialized Unfold remains costly. Optional patch-wise inference reduces working memory further.

On one AMD EPYC 9355 core (batch 1, one PyTorch/OMP/MKL thread), complete 2K inference yields $1.268\times$ and $1.402\times$ speedups for SRResNet-Ours and SwinIR-Ours over their dense backbones, and SRResNet-Ours has latency comparable to ClassSR and ARM. CNN timing includes patch stitching, and SwinIR uses whole-image inference; loading and disk I/O are excluded. Benefits remain implementation- and hardware-dependent.}

\subsection{Comparative Results on Low-Resolution Benchmarks}
\label{sec:low_test}

The acceleration benefits of super-resolution become most evident for
high-resolution inputs. To nevertheless verify the effectiveness of
our frequency-guided policy from multiple aspects, we select three
classical low-resolution benchmarks---Set5, Set14, and Urban100---and
conduct experiments on $\times2$, $\times3$, and $\times4$ SR. All
models are trained on DIV2K using identical settings. We take the
CNN-based method CARN and the Transformer-based method SwinIR as
examples. As shown in \revise{Supplementary Table~S7}, our strategy
maintains comparable reconstruction quality with reduced computation
across different upscaling factors and backbones, demonstrating its
generality across architecture types and upscaling factors.

\subsection{Robustness under Degradation Shift}\label{sec:unknown_deg}

\begin{figure*}[!htbp]\footnotesize
	\centering
	\includegraphics[width=\linewidth]{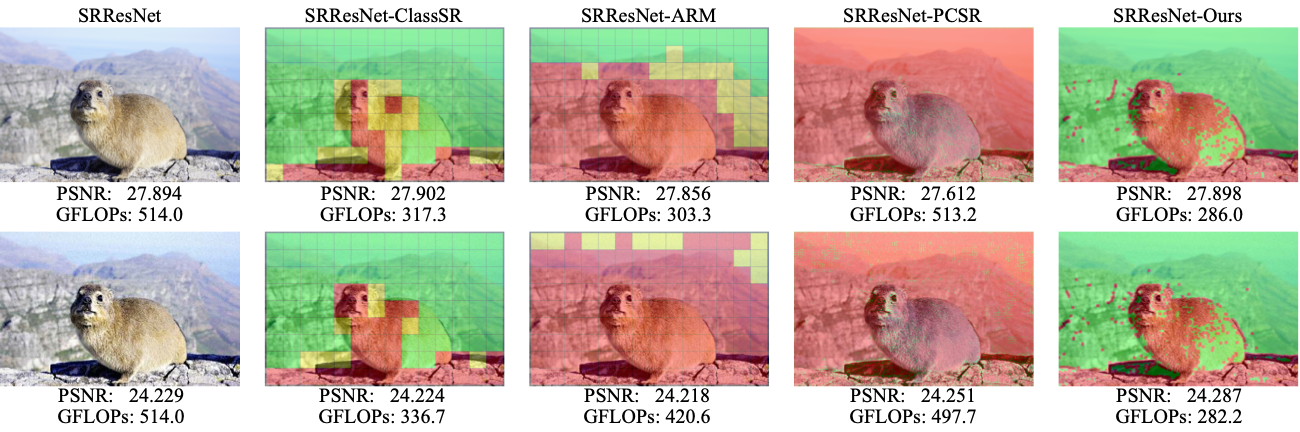}   %\\
	\caption{
		Illustration of the $\times 4$ SR results of the image ``1244" from Test2K under different degradations. The first row presents the results when the degradations are aligned with those used during training~\cite{kong2021classsr}. The second row displays the results employing a more extensive and complex set of random degradations~\cite{zhang2021designing}. 
	}
	\label{fig:unknown_deg}
\end{figure*}

To evaluate the stability of computation allocation under degradation
shift, we employ the random degradation pipeline
in~\cite{zhang2021designing}, which introduces additional degradations
such as noise and compression.
Fig.~\ref{fig:unknown_deg} visualizes the behavior of different
adaptive methods under the degradation shift. \revise{Adaptive} methods
may assign more regions to computationally expensive paths when
degradation artifacts are interpreted as difficult image content.
This behavior is particularly evident for ARM.
Using SRResNet and SwinIR as examples, we provide quantitative results
on Test2K in Table~\ref{table:gen}. Under \revise{mixed degradation}, the
relative FLOPs changes of ClassSR, ARM, and PCSR are $+5.2\%$,
$+21.1\%$, and $+6.1\%$, respectively. In contrast, the proposed policy
results in smaller changes of $+3.0\%$ for SRResNet and $+3.6\%$ for
SwinIR, while maintaining comparable reconstruction quality. These
results indicate that the proposed non-parametric policy provides more
stable computation allocation under the evaluated degradation shift.

\revise{Table~\ref{table:gen} adds individual noise, JPEG, and blur conditions with fixed inference settings. The evaluation concerns the joint quality--computation trade-off, rather than invariant reconstruction quality. Noise and compression can trigger nonstructural frequency responses; our allocation also changes but drifts less than the compared SRResNet-based adaptive methods.}

\begin{table}[!htbp]
\footnotesize
\setlength{\tabcolsep}{3pt}
\centering
\caption{\revise{Test2K reconstruction quality and computation under individual and mixed degradation shifts. Each cell lists PSNR / FLOPs; parentheses give the signed FLOPs change from bicubic input for the same method. Noise uses $\sigma_n=20$, JPEG uses $Q=50$, and blur uses $\sigma_b=2.0$. Inference settings remain fixed across conditions.}}
\label{table:gen}
\resizebox{\linewidth}{!}{%
\begin{tabular}{l|c|c|c|c|c}
\hline
\hline
Method
& Bicubic
& Gaussian Noise
& JPEG Compression
& Gaussian Blur
& Mixed Degradation \\
\hline

SRResNet
& 26.223 / 502.9G
& 24.018 / 502.9G (0.0\%)
& 25.396 / 502.9G (0.0\%)
& 26.003 / 502.9G (0.0\%)
& 23.014 / 502.9G (0.0\%) \\

SRResNet-ClassSR
& 26.212 / 445.2G
& 24.013 / 458.7G (+3.0\%)
& 25.395 / 451.2G (+1.3\%)
& 26.002 / 448.9G (+0.8\%)
& 23.010 / 468.5G (+5.2\%) \\

SRResNet-ARM
& 26.215 / 394.8G
& 24.013 / 472.5G (+19.7\%)
& 25.393 / 423.1G (+7.2\%)
& 26.001 / 410.8G (+4.1\%)
& 23.009 / 478.2G (+21.1\%) \\

SRResNet-PCSR
& 26.216 / 388.2G
& 24.016 / 405.7G (+4.5\%)
& 25.395 / 398.1G (+2.6\%)
& 26.002 / 392.2G (+1.0\%)
& 23.020 / 411.8G (+6.1\%) \\

SRResNet-Ours
& 26.223 / 349.6G
& 24.016 / 358.2G (+2.5\%)
& 25.398 / 353.4G (+1.1\%)
& 26.006 / 352.0G (+0.7\%)
& 23.029 / 360.1G (+3.0\%) \\

\hline

SwinIR
& 26.323 / 102.6G
& 24.194 / 102.6G (0.0\%)
& 25.427 / 102.6G (0.0\%)
& 26.008 / 102.6G (0.0\%)
& 23.216 / 102.6G (0.0\%) \\

SwinIR-Ours
& 26.347 / 78.4G
& 24.210 / 80.1G (+2.2\%)
& 25.449 / 79.1G (+0.9\%)
& 26.010 / 78.9G (+0.6\%)
& 23.228 / 81.2G (+3.6\%) \\

\hline
\hline
\end{tabular}
}
\end{table}

\begin{figure*}[!htbp]
    \centering
    \includegraphics[width=\linewidth]{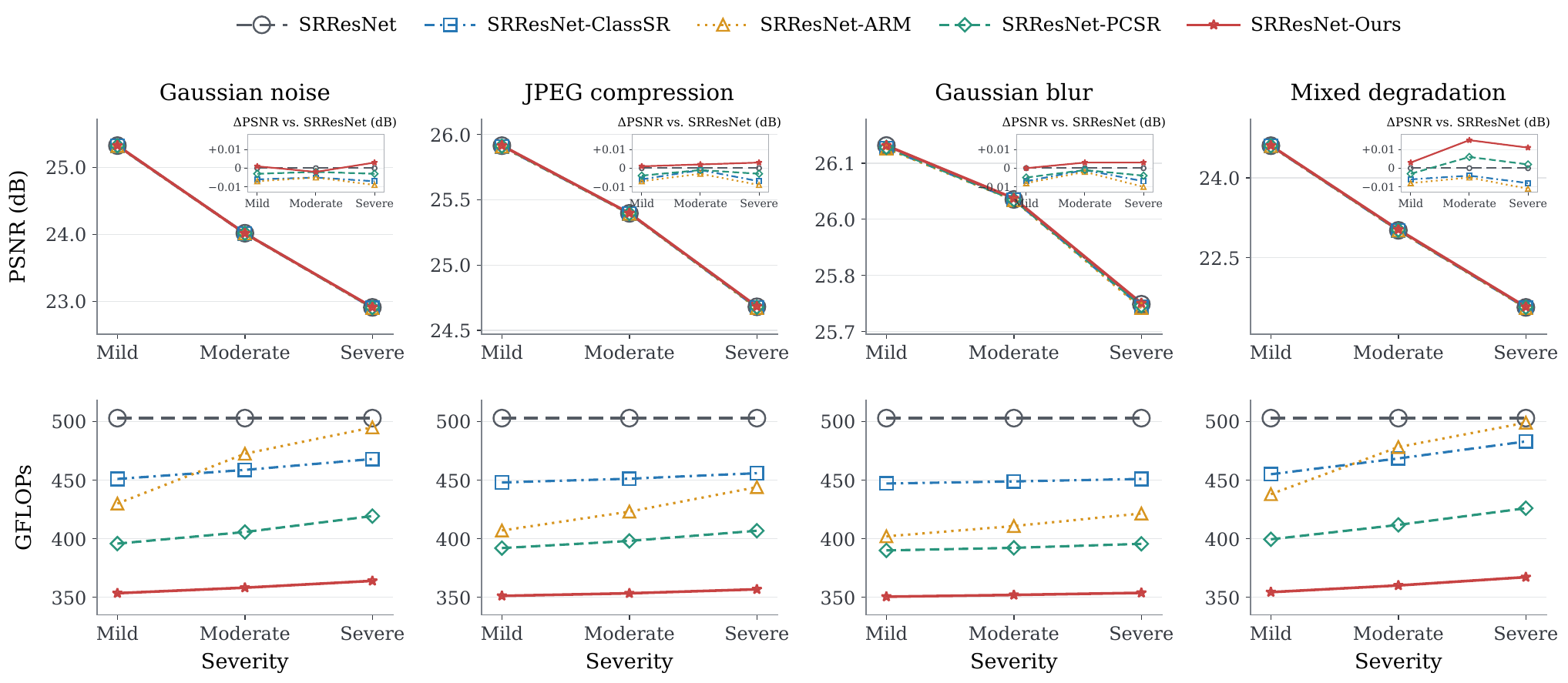}
    \caption{\revise{Severity evaluation on Test2K with fixed inference settings. Columns: noise, JPEG, blur, and mixed degradation. Rows: PSNR (insets: difference from dense SRResNet) and GFLOPs. Mixed conditions combine corresponding component strengths.}}
    \label{fig:degradation_severity}
\end{figure*}

\revise{Fig.~\ref{fig:degradation_severity} varies noise ($\sigma_n=10,20,30$), JPEG ($Q=70,50,30$), and blur ($\sigma_b=1,2,3$). Mixed conditions combine corresponding strengths through the same pipeline~\cite{zhang2021designing}. All routing settings remain fixed. Quality decreases with severity, while our method retains comparable quality, the lowest adaptive FLOPs, and smaller computation drift; noise/JPEG cause larger changes than blur. Dense SRResNet has constant FLOPs by design. These controlled synthetic results do not establish universal real-world robustness.}

%-------------------------------------------------------------------------
\subsection{Ablation Study}\label{sec:abl}

\subsubsection{Propagation Range}\label{sec:kernel}
\begin{figure}[!htbp]\footnotesize
	\centering
	\includegraphics[width=\linewidth]{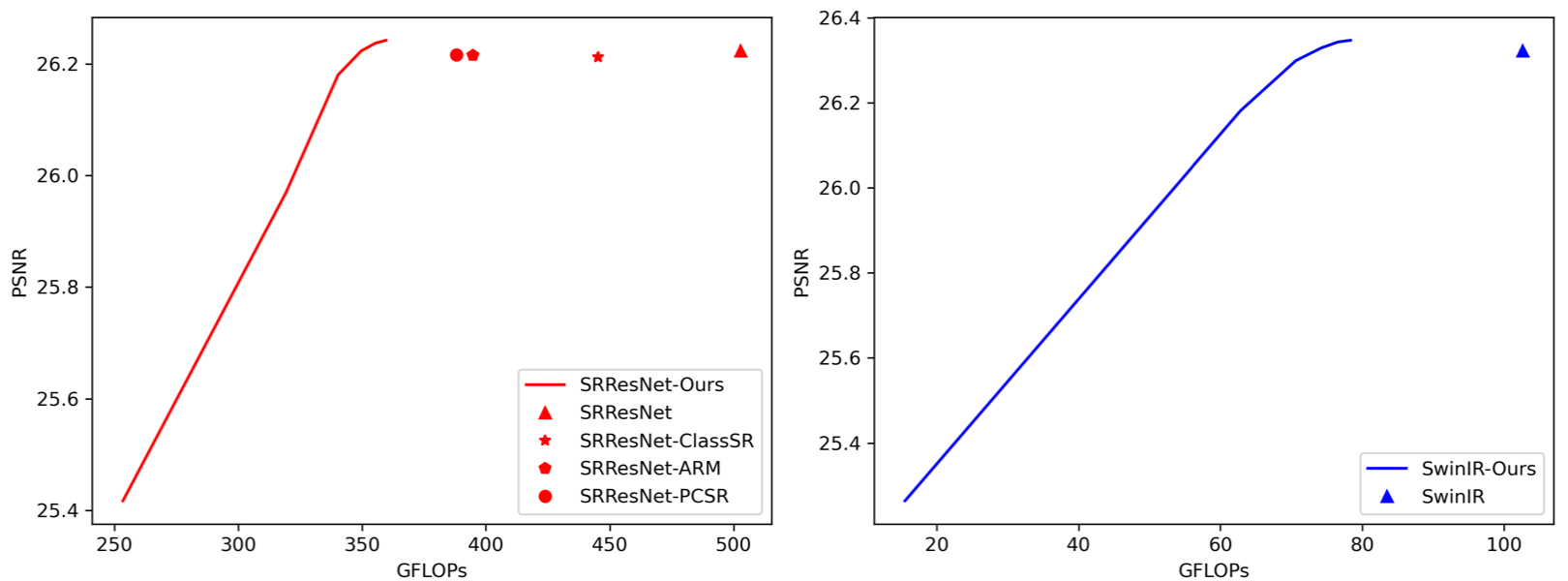}   %\\
	\caption{
		The impact of propagation range on various backbones.
		Expanding the propagation range improves reconstruction quality
		while increasing computation.
	}
	\label{fig:ks}
\end{figure}

Our method supports computation adjustment through the propagation
range $K$ without requiring retraining. As shown in
Fig.~\ref{fig:ks}, we analyze the effects of varying $K$ from 1 to 11
on Test2K for both the CNN-based SRResNet and the Transformer-based
SwinIR.
The body of SwinIR accounts for a substantially larger proportion of
the overall computation (99\%) than that of SRResNet (47\%), making
its computation more sensitive to propagation-range adjustment.
Despite this architectural difference, our method obtains results
comparable to the corresponding backbones while achieving
computational savings of 30.5\% and 31\% for SRResNet and SwinIR,
respectively\revise{, at representative ablation operating points of $K=7$ for SRResNet and $(K,\sigma)=(5,0.5)$ for SwinIR}. \revise{The validation-selected settings used in the main comparison are specified in Section~\ref{sec:implementation_details}.}

The input-dependent policy and propagation-range adjustment operate at
different levels. The adaptive policy determines \emph{where}
computation is needed according to the frequency response of each
input image, whereas $K$ controls \emph{how much} computation is
allocated by changing the spatial extent of the active regions.
Adjusting $K$ therefore changes the overall sparsity level without
replacing the underlying input-dependent decision process.

\subsubsection{Window Activation Threshold}
Unlike CNN-based approaches, window-based Transformers jointly process
the tokens within each attention window. This architectural
distinction introduces two hyperparameters governing computational
efficiency: the propagation range $K$ and the window activation
threshold $\sigma$. With $K$ fixed to 5, we evaluate the sensitivity
to $\sigma$ in terms of reconstruction quality and computation. As
shown in Table~\ref{table:thres}, $\sigma=0.5$ achieves a 31--50\%
FLOPs reduction while maintaining reconstruction quality
($\Delta$PSNR $<0.16$ dB) across the benchmark datasets.
\begin{table}[!htbp]\footnotesize  %[t]  %
	\centering
	\caption{The impact of the window activation threshold $\sigma$ on
		SwinIR, evaluated using the same trained checkpoint with $K=5$.}
	\label{table:thres}
%	\resizebox{\linewidth}{!}{
		\begin{tabular}{c|cc|cc|cc}
			
			\hline	
			\multirow{2}{*}{$\sigma$}  &  \multicolumn{2}{c|}{Test2K}  &  \multicolumn{2}{c|}{Test4K}   &  \multicolumn{2}{c}{Test8K}\\
			\cline{2-7}
			&   PSNR   &  GFLOPs   &   PSNR  &  GFLOPs  &  PSNR   &  GFLOPs  \\
			\hline	
			1.0 &  25.264 & 15.5 & 26.469  & 61.3  &  32.267  & 350.6  \\
			0.7 & 26.218  & 66.0   & 27.640  & 237.8  & 33.513  &  1097.3    \\
			0.5 & 26.299  & 70.6   & 27.746  & 256.8  & 33.611  & 1207.6  \\
			0.3 & 26.331  & 73.9  & 27.787   & 270.8    & 33.645  & 1293.1    \\
			0.0 & 26.356  & 102.6  & 27.828   & 415.0  & 33.686  & 2399.8 \\
			\hline	
			
		\end{tabular}
%	}
\end{table}

\revise{Supplementary Figs.~S3 and~S4 provide visual examples of test-time propagation and window-threshold adjustment using unchanged checkpoints.}

\subsubsection{Mask Generation}\label{sec:mask}
We additionally analyze \revise{eight} strategies for generating execution masks: 1) a random mask with matched sparsity; 2) normalizing the high-frequency response and applying a fixed threshold of 0.5; 3) comparing the high-frequency response with its per-image median\revise{ (50th percentile)}; \revise{4) Gaussian high-pass response with Otsu thresholding~\cite{otsu1979threshold}; 5--7) LoG, local variance, or local entropy with two-cluster partitioning; and 8)} applying the proposed per-image two-cluster partitioning. For the random baseline, we randomly select exactly the same number of active locations as in the mask generated by our method for each input image. Therefore, the random and proposed masks have identical per-image active ratios, while differing only in their spatial allocation. We train the random-mask variant three times using independent random seeds and report the mean results. To isolate the effect of the mask-generation strategy, we use CARN and set $K=3$ for all variants to obtain comparable computation. Each variant is trained independently using its corresponding mask-generation strategy. 
Note that this controlled ablation setting differs from the $K=5$ setting used for CARN in the main comparison; therefore, the PSNR and FLOPs reported in Table~\ref{table:mask} are not expected to match those in Tables~\ref{table:sota} and~\ref{table:time}. The results are shown in Table~\ref{table:mask}. The proposed two-cluster strategy adapts the policy to each image and achieves \revise{the best observed quality--computation trade-off among the evaluated variants}.

\revise{Alternative indicators retain the same two-cluster partitioning and propagation. We compare their naturally resulting quality--computation trade-offs without indicator-specific thresholds to force identical FLOPs. On Test8K, LoG, local variance, and entropy all use more computation but achieve lower PSNR than our policy. Matched-sparsity random masks perform worse, supporting the importance of spatial placement.

Otsu and one-dimensional two-cluster partitioning are closely related variance-based rules. Their small observed difference is empirical rather than evidence of fundamental superiority. Normalization preserves optimal assignments (Section~\ref{sec:policy}); $K$ and $\sigma$ sensitivity are examined separately.}

\begin{figure}[!htbp]\footnotesize
	\centering
	\includegraphics[width=0.7\linewidth]{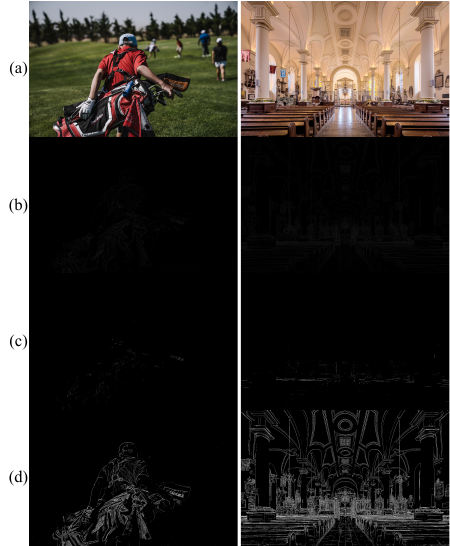} 
	\caption{
		(a) LR images, (b) normalized Gaussian high-pass magnitude
		responses, (c) execution masks generated using a fixed threshold
		of 0.5, and (d) execution masks generated using the proposed
		two-cluster partitioning. Active locations are shown in white.
	}
	\label{fig:mask_gen}
\end{figure}
\begin{table}[!htbp]
\footnotesize
\setlength{\tabcolsep}{4.5pt}
\centering
\caption{\revise{Mask-generation ablation with independently trained CARN variants ($K=3$). Indicators share clustering and propagation. Random masks match per-image active ratios (mean of three seeds); median denotes the 50th percentile.}}
\label{table:mask}
\resizebox{\linewidth}{!}{
\begin{tabular}{l|cc|cc|cc}
\hline
\hline
\multirow{2}{*}{Mask generation}
& \multicolumn{2}{c|}{Test2K}
& \multicolumn{2}{c|}{Test4K}
& \multicolumn{2}{c}{Test8K} \\
& PSNR & GFLOPs
& PSNR & GFLOPs
& PSNR & GFLOPs \\
\hline

Random mask (matched sparsity)
& 25.023 & 61.83
& 26.152 & 244.5
& 31.893 & 1335.8 \\

Fixed threshold
& 25.552 & 60.07
& 26.791 & 237.2
& 32.593 & 1299.2 \\

Per-image median (50th percentile)
& 25.550 & 61.58
& 26.814 & 251.4
& 32.601 & 1445.8 \\

Gaussian high-pass + Otsu thresholding
& 25.876 & 61.76
& 27.300 & 245.1
& 32.988 & 1333.2 \\

\hline

LoG + two-cluster
& 25.784 & 62.09
& 27.005 & 249.2
& 32.796 & 1421.0 \\

Local variance + two-cluster
& 25.573 & 63.46
& 26.883 & 259.2
& 32.737 & 1511.1 \\

Local entropy + two-cluster
& 25.556 & 63.20
& 26.839 & 254.4
& 32.681 & 1499.0 \\

Gaussian high-pass + two-cluster (Ours)
& 25.927 & 61.83
& 27.307 & 244.5
& 33.162 & 1335.8 \\

\hline
\hline
\end{tabular}
}
\end{table}
Fig.~\ref{fig:mask_gen} qualitatively compares the masks generated by
the fixed-threshold strategy and the proposed two-cluster strategy.
Specifically, Fig.~\ref{fig:mask_gen}(c) applies a fixed threshold of
0.5 to the normalized Gaussian high-pass magnitude response in
Fig.~\ref{fig:mask_gen}(b), regardless of the image-dependent response
distribution. 
In contrast, the proposed strategy in
Fig.~\ref{fig:mask_gen}(d) adapts the partition boundary to the
response distribution of each image, producing active regions that are
more closely aligned with content-responsive structures. The
corresponding quantitative quality--computation comparison is reported
in Table~\ref{table:mask}.

\subsubsection{\revise{Training-Time Propagation Schedule}}
\label{sec:training_schedule}

\revise{Supplementary Table~S6 compares independently trained SRResNet variants with unchanged optimization: fixed $K=8$ (the range midpoint), uniform sampling from $\{5,\ldots,11\}$, and progressive $11\rightarrow5$. All use validation-selected $K=7$ at inference, yielding identical masks and FLOPs. Gains of only 0.002--0.006 dB do not establish meaningful schedule superiority; progressive training is retained as a simple broad-to-selective curriculum.}

\subsection{\revise{Failure-Case Analysis}}
\label{sec:failure_cases}

\begin{figure*}[!htbp]
    \centering
    \includegraphics[width=\linewidth]{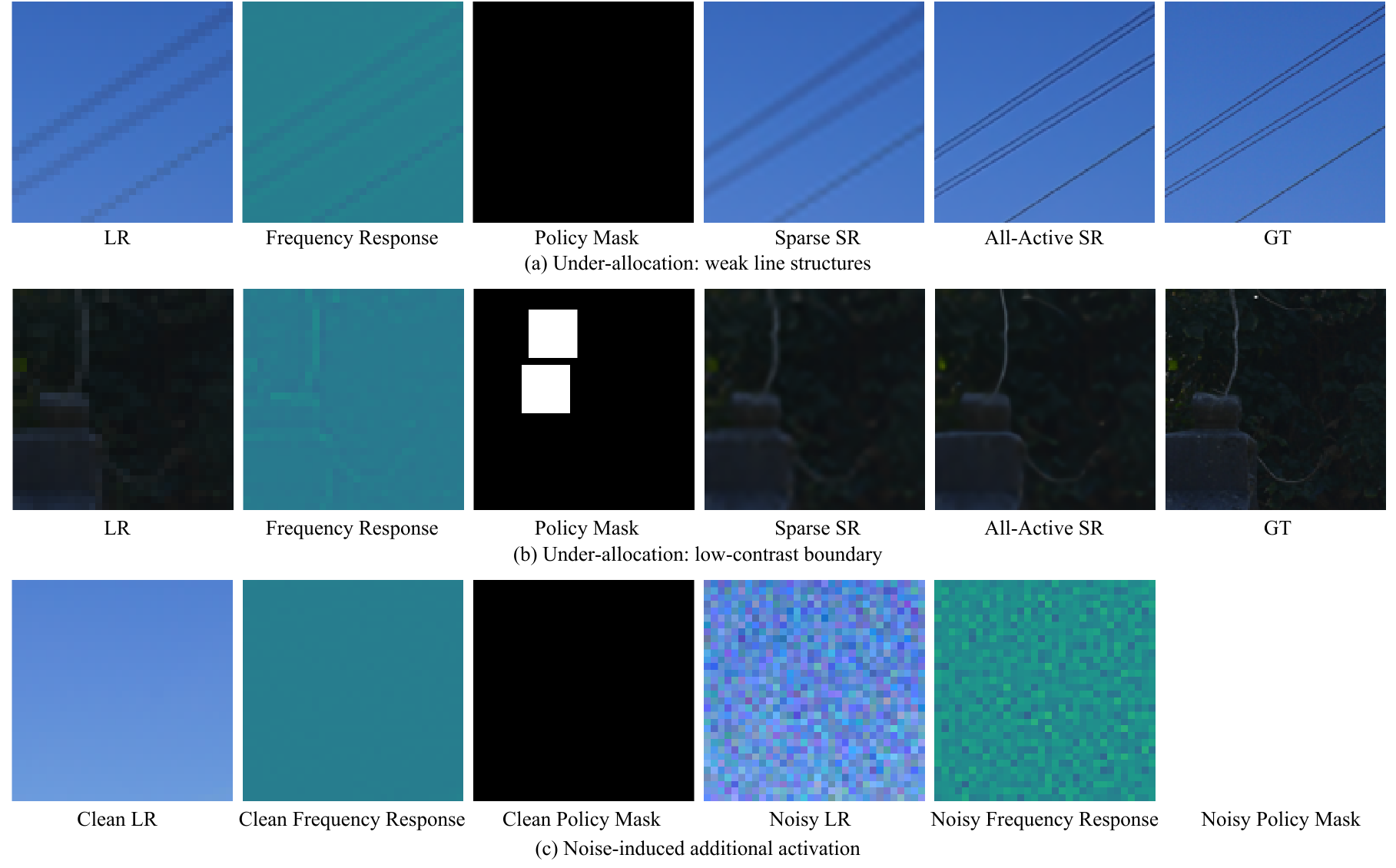}
    \caption{\revise{Policy failure cases. (a,b) Weak lines and a low-contrast boundary receive insufficient computation; all-active inference improves reconstruction. (c) Noise ($\sigma_n=20$) raises the within-image normalized score and changes activation from 0 to 1, illustrating possible over-allocation.}}
    \label{fig:failure_cases}
\end{figure*}

\revise{ARM~\cite{chen2022arm} reports Spearman correlation 0.85 between PSNR and negative LR-patch Laplacian score; FAD~\cite{xie2021learning} also uses frequency-based allocation. Building on this prior, we examine failures of our deployed policy rather than treating its score as an exact difficulty estimator.

With one SRResNet-Ours checkpoint ($K=7$), we compare sparse and all-active inference on 9,192 non-overlapping $32\times32$ LR patches from 100 Test2K images. Each patch has a mean per-image normalized frequency score, final active ratio, and error difference $\Delta E=\mathrm{MSE}_{\mathrm{sparse}}-\mathrm{MSE}_{\mathrm{all\text{-}active}}$. Pre-specified screening selects the bottom 30\% in score and active ratio and top 20\% in positive $\Delta E$, giving 154 candidates.

Fig.~\ref{fig:failure_cases}(a,b) shows under-allocation to weak lines and a low-contrast boundary: active ratios are 0 and 0.096; all-active PSNR improves from 30.98 to 37.71 dB and 35.59 to 36.68 dB. In (c), noise ($\sigma_n=20$) changes the within-image normalized score from 0.4250 to 0.4919 and activation from 0 to 1. This illustrates possible over-allocation to artifacts without establishing that all additional computation is unnecessary. The analysis characterizes failure modes, not a general score--benefit correlation.}

\section{Limitations and Future Work}
Although the proposed framework achieves favorable
quality--computation trade-offs across different SR backbones, several
limitations remain. First, the current acceleration is primarily
applied to the repeated transformations in the network body.
Consequently, the maximum attainable acceleration depends on the
proportion of total computation contributed by the conditionally
executed body. For architectures in which the reconstruction and
upsampling modules account for a relatively large fraction of the
overall computation, such as SRResNet, these densely executed modules
impose a fixed lower bound on the total computational cost.

Second, the proposed policy uses image-frequency responses as a simple
and router-free indicator of spatial reconstruction demand. Although
this prior provides more stable computation allocation than the
evaluated \revise{adaptive} methods under degradation shift, noise and
compression artifacts may still introduce additional frequency
responses and activate \revise{additional} locations. \revise{Weak lines and low-contrast boundaries can also be under-allocated, as shown in Section~\ref{sec:failure_cases}. The frequency score is a routing prior rather than a calibrated measure of local computation demand. The controlled synthetic degradation experiments do not establish universal reconstruction robustness.} The current policy is
therefore most effective when frequency-responsive structures are
closely associated with regions requiring greater reconstruction
capacity.

Third, the practical runtime benefit of active-site computation depends
on the efficiency of local-descriptor gathering, feature scattering,
and active-window batching on the target hardware. As a result, FLOPs
reduction may not translate proportionally into runtime reduction for
all architectures or deployment platforms. In addition, dense tensor
operations are retained during training to support efficient
mini-batch optimization, and the current framework mainly targets
inference acceleration. \revise{The current CNN implementation explicitly materializes \texttt{Unfold}; a measured reduction in net peak memory does not make this temporary representation cost-free. Profiling exposes substantial data-reorganization costs, and GPU and CPU measurements characterize the evaluated implementations.}

Future work will investigate adaptive execution for the reconstruction
and upsampling stages to further reduce the remaining dense
computation. It is also promising to develop hardware-optimized sparse
operators and to incorporate complementary content cues while
preserving the router-free and input-adaptive nature of the proposed
policy.

\section{Conclusion}
We presented a frequency-guided adaptive computation framework for
efficient single-image super-resolution. The proposed framework
constructs an input-dependent execution policy directly from the
frequency response of each LR image, without introducing a learned
routing module or auxiliary routing supervision. Based on this policy,
CNN backbones selectively execute body transformations at active
spatial locations, whereas window-based Transformers perform
conditional execution at the window level. The framework retains a
regular masked computation graph during training and skips inactive
locations or windows during inference.

Experiments across multiple CNN and Transformer backbones demonstrate
that the proposed method reduces computational cost and measured
runtime while maintaining comparable reconstruction and perceptual
quality. \revise{Recent-method comparisons, independent-seed reconstruction statistics, and repeated runtime, memory, CPU, and component-level measurements further characterize these trade-offs.} The propagation range and window activation threshold further
allow the quality--computation trade-off to be adjusted at test time
without retraining or modifying the learned network parameters. These
results show that classical image priors can provide a simple and
effective basis for input-adaptive computation when properly integrated
with modern image restoration architectures.

\section*{Funding}
This work was supported in part by the National Natural Science Foundation of China (62576241, 62172127, and U22B2035).

\section*{CRediT authorship contribution statement}
\textbf{Wei Shang:} Writing – original draft, Visualization, Validation, Methodology, Investigation, Software. \textbf{Wanying Zhang:} Visualization, Validation.  \textbf{Wenwen Li:} Conceptualization. \textbf{Xiaohe Wu:} Conceptualization. \textbf{Huaming Wu:} Resources. \textbf{Dongwei Ren:} Writing – review \& editing, Supervision. 

\section*{Declaration of competing interest}
The authors declare that they have no known competing financial interests or personal relationships that could have appeared to influence the work reported in this paper.

\section*{Data availability}

The datasets used in this study are publicly available from the sources cited in Section~\ref{sec:dataset}. No new dataset was created for this work. The source code and trained models will be made available at \url{https://github.com/shangwei5/AMSR} upon publication.

\clearpage
\bibliographystyle{elsarticle-num} 
\bibliography{sn-bibliography}

\end{document}